\documentclass[letterpaper, 10 pt, conference]{ieeeconf}

\IEEEoverridecommandlockouts                              

\usepackage{times}
\usepackage{epsfig}
\usepackage{graphicx}
\usepackage{amsmath}
\usepackage{amssymb}
\usepackage{algorithm,xspace}
\usepackage{algcompatible}
\usepackage{multirow}
\usepackage{xcolor}
\usepackage{subfigure}

\def\etal{\emph{et al}.}

\def\eg{\emph{e.g.}}
\def\ie{\emph{i.e.}}

\renewcommand{\mathbf}{\boldsymbol}
\renewcommand{\O}{\mathcal{O}}
\newcommand{\T}{\mathcal{T}}

\newcommand{\N}{\mathcal{N}}

\newcommand{\x}{\mathbf{x}}
\newcommand{\y}{\mathbf{y}}
\newcommand{\z}{\mathbf{z}}
\renewcommand{\v}{\mathbf{v}}

\newcommand{\nop}[1]{}

\newtheorem{example}{Example}
\newtheorem{problem}{Problem}

\title{\LARGE \bf
Data-driven Distributed State Estimation and Behavior Modeling in Sensor Networks}
\author{Rui Yu$^{1}$, Zhenyuan Yuan$^{2}$, Minghui Zhu$^{2}$, Zihan Zhou$^{1}$
\thanks{$^{1}$R. Yu and Z. Zhou are with College of Information Sciences and Technology, Pennsylvania State University, University Park, PA 16802, USA {\tt\small \{rzy54, zuz22\}@psu.edu}}%
\thanks{$^{2}$Z. Yuan and M. Zhu are with School of Electrical Engineering and Computer Science, Pennsylvania State University, University Park, PA 16802, USA {\tt\small \{zqy5086, muz16\}@psu.edu}}%
\thanks{This work is supported in part by a seed grant from Penn State Institute for Computational and Data Sciences. Z. Yuan and M. Zhu were supported by NSF grants ECCS-1710859 and CNS-1830390.}%
}

\begin{document}

\maketitle
\thispagestyle{empty}
\pagestyle{empty}

\begin{abstract}

Nowadays, the prevalence of sensor networks has enabled tracking of the states of dynamic objects for a wide spectrum of applications from autonomous driving to environmental monitoring and urban planning. However, tracking real-world objects often faces two key challenges: First, due to the limitation of individual sensors, state estimation needs to be solved in a collaborative and distributed manner. Second, the objects' movement behavior model is unknown, and needs to be learned using sensor observations. In this work, for the first time, we formally formulate the problem of simultaneous state estimation and behavior learning in a sensor network. We then propose a simple yet effective solution to this new problem by extending the Gaussian process-based Bayes filters (GP-BayesFilters) to an online, distributed setting. The effectiveness of the proposed method is evaluated on tracking objects with unknown movement behaviors using both synthetic data and data collected from a multi-robot platform.
\end{abstract}

\section{Introduction}
Estimating the states of dynamic phenomena has long been a central problem in robotics with numerous real world applications. Taking autonomous driving as an example, for the autonomous cars to safely and efficiently navigate in a cluttered environment (\eg, city traffic), it is critical to estimate the movements of other traffic agents (\eg, pedestrians, bicyclists, and human-driven vehicles) and predict their future trajectories. Recent advances in sensing technologies have enabled the estimation and understanding of traffic movements using streaming data collected from various sensors such as on-board cameras, radar, and LiDAR, as well as surveillance cameras.

However, current state estimation methods still face two major challenges in such real scenarios. \emph{First}, the capability of individual sensor is limited by its perception range (\eg, line-of-sight and field-of-view) and also subject to noises and errors. Thus, it is often critical for the sensors to fuse information of others (\eg, via vehicle-to-vehicle and vehicle-to-infrastructure communication networks) for accurate estimation. In other words, state estimation needs to be solved in a collaborative and distributed manner. \emph{Second}, the movement behavior of traffic agents is highly non-linear, influenced by both their internal goals and interactions with other agents. Usually the internal goals and interactions are unknown, thus the movement behavior needs to be learned online using sensor observations.

Rather surprisingly, the two challenges mentioned above have never been studied together. Therefore, the \textbf{first contribution of this work} is that we formally formulate the problem of simultaneous state estimation and behavior modeling in a sensor network, and discuss its relation to existing work in the literature. Here, we emphasize an \emph{online} learning setting, \ie, we do not assume access to any training data with noise-free movement states, which are often difficult to obtain in real applications. Instead, the behavior model must be learned and updated using noisy sensor observations.

The \textbf{second contribution of this work} is a simple yet effective method to tackle this new problem. Our method is based on Bayes filtering with Gaussian process models (GP-BayesFilters)~\cite{ko2009gp}. While traditional Bayes filters assume known (\eg, parametric) motion and observation models, GP-BayesFilters use Gaussian process regression~\cite{RasmussenW06} to learn such models from data. We choose GP models over other data-driven models (\eg, neural networks) based on two considerations. \emph{First}, GP as a non-parametric model can learn from small-scale data, which is a competitive advantage in online learning. \emph{Second}, GP models generate state-dependent uncertainty estimates that consider both noise and regression uncertainty due to limited training data, allowing them to be readily incorporated into various forms of Bayes filters (\eg, extended Kalman filters). 

In this work, we extend the GP-based Bayes filters in two aspects: (i) We apply it to distributed state estimation by using the consensus algorithms~\cite{Olfati-SaberFM07}, where each sensor node can share information with neighbors and update the state by fusing information received from its neighbors; (ii) We show how the GP motion and observation models can be learned from noisy and partial observations in an online fashion, without requiring training data containing sequences of ground truth states.

The effectiveness of the proposed method is verified by experiments on both synthetic data and data collected from a multi-robot platform. We show that our data-driven GP models outperform pre-defined (\ie, linear) models in prediction and tracking of traffic agents with unknown movement behavior. And the GP models learned by fusing information from multiple sensors via consensus are better than those learned by each sensor individually. In the experiment on robot platform, we show the advantage of data-driven models over parametric models in handling un-modeled system dynamics. Finally, we demonstrate that, compared to more complicated models (\ie, deep neural networks), the GP model is a better choice for learning from small-scale noisy data in an online fashion.

\section{Related Work}

\subsection{State Estimation with Gaussian Process Models}
Gaussian process state space models (GPSSMs) are often used in state estimation problems for filtering~\cite{ko2009gp,DeisenrothHH09} or smoothing~\cite{DeisenrothTHHR12}. By describing the unknown transition and observation functions using GP models, Ko~\etal~\cite{ko2009gp} and Deisenroth~\etal~\cite{DeisenrothHH09} derive different forms of GP-based Bayes filters including GP-EKF, GP-UKF, GP-PF, and GP-ADF. Since exact GP regression is challenging for large datasets due to $O(n^3)$ time complexity, \cite{PanYTB17} proposes sparse spectrum Gaussian process (SSGP), which maps the input to a low-dimensional feature space for fast inference. The original GP-BayesFilters also require ground truth states to train the GP models. To alleviate this issue, \cite{KoF11} first determines the latent states from observations by introducing a Gaussian Process prior over the latent space. Our work further extends the GPSSMs in two different ways, by (i) introducing consensus algorithms to the GP-BayesFilters framework for distributed state estimation, and (ii) learning the GP models online, where the states are estimated from observations using Kalman filters.

\subsection{Distributed State Estimation}
Consensus~\cite{Olfati-SaberFM07, ZhuM10} is a major methodology for distributed state estimation over sensor networks. The basic idea is that each network node runs a local filter, and information is spread through consensus of neighboring nodes. Several consensus algorithms have been developed, including consensus on estimation (CE)~\cite{Olfati-Saber07a}, consensus on measurement (CM)~\cite{olfati2005distributed}, and consensus on information (CI)~\cite{battistelli2011information}. \cite{BattistelliCMFG15} combines several algorithms into a general class of consensus filters named Hybrid CMCI. The stability of Hybrid CMCI is further proven in~\cite{BattistelliC16}. 
The consensus algorithm we adopt falls in this general class.
But unlike existing consensus filters which assume known state transition and observation models, we study data-driven approaches to learn such models.

\subsection{Multi-agent Behavior Modeling}
Our work is also related to the line of research which attempts to learn the movement behavior of traffic agents. To describe human movement dynamics and interactions, several methods~\cite{PellegriniESG09, LuberSTA10, YamaguchiBOB11} use the social force model~\cite{Helbing95}. In these works, the model parameters are learned offline using manually annotated trajectories. \cite{TamuraLHCBYA12} identifies the model parameters through a series of carefully designed observation experiments.
\cite{EllisS009} learns pedestrian motion patterns with Gaussian Processes models.  
Recently, pedestrian behaviors are modeled using deep networks such as LSTM~\cite{AlahiGRRLS16} and GAN~\cite{GuptaJFSA18}.
However, these methods do not consider distributed state estimation, and rely on ground truth states to learn the models.

\section{Problem Formulation}
\label{sec:pro}
In this work, we address the problem of collaborative tracking and behavior modeling of dynamic objects via a sensor network. Specifically, we are interested in a system composed of $n$ sensors. The communication network connecting the sensors is represented by an undirected graph $G=(V,E)$ where vertices $V = \{1, \ldots, n\}$ correspond to the sensors and an edge $(i,i')\in E$ indicates that sensors $i$ and $i'$ can communicate. We denote the neighbors of sensor $i$ as $N_i = \{i' | (i,i') \in E, i' \in V \setminus \{i\}\}$.

Consider an environment that consists of $m$ targets, whose states are represented by $\{\y_1, \ldots, \y_m\}$. We assume each target $\y_j$ evolves according to the following \emph{target motion model}:
\begin{equation}
\y_j^{t+1}=f_j^t(\y_1^t, \ldots, \y_m^t) + \eta_j^t,
\label{eq001}
\end{equation}
where $\eta_j^t \sim \N(0, \Sigma_{\eta_j})$ is the process noise.

At any time $t$, sensor $i$ can obtain state measurements of a subset of the targets.
Let $\O_j^t \subseteq \{1, \ldots, n\}$ be the set of sensors which can observe target $j$ at time $t$, we assume a general \emph{sensor observation model}:
\begin{equation} \z_{ij}^t = h_{ij}^t(\y_j^t)+\nu_{ij}^t, \quad \forall i \in \O_j^t, \label{eq002}
\end{equation}
where $\nu_{ij}^t \sim \N(0, \Sigma_{\nu_{ij}})$ is the measurement noise.

\begin{problem} Suppose the system function $f_j$ and output function $h_{ij}$ are time-invariant but remain unknown. The task of each sensor $i$ is to construct estimators of the target states $\{\hat{\y}_j^t\}$ in a distributed fashion, \ie, using information only from its neighbors and the measurements $\{\z_{ij}^t\}$.
\label{prob1}
\end{problem}

In the following, we take the practical task of estimating and modeling human movements in traffic scenes as an example to illustrate our problem formulation.  

\begin{example}\label{sec:pro:example}
Consider the scenario of tracking multiple traffic agents (\eg, pedestrians). Each agent state $\y_j$ consists of the position $\x_j$ and velocity $\v_j$. The dynamics of the system can be described as follows:
\begin{eqnarray}
\x_j^{t+1} & = & \x_j^t + \v_j^t \cdot \Delta t + \eta_{j,\x}^t \nonumber \\ 
\v_j^{t+1} & = & \v_j^t + g(\y_1^t, \ldots, \y_m^t; \theta_j) \cdot \Delta t + \eta_{j,\v}^t
\label{eq:motion_eq}
\end{eqnarray}
where $\Delta t$ is the sampling period and $g$ is defined below.

We assume the motion of each agent follows a simplified version of social force model (SFM)~\cite{Helbing95}, a popular formula for modeling human movement behavior in traffic scenes. According to SFM, each agent wants to keep its desired velocity, but is influenced by the other agents:
\begin{align}
\frac{d \v_j^t}{d t} & = g(\y_1^t, \ldots, \y_m^t; \theta_j) \nonumber \\
& = \frac{1}{\tau_j} (\v_j^* - \v_j^t) + \sum_{k=1}^m \alpha_j e^{\frac{-\|\x_j^t - \x_k^t\|}{\beta_j}} \frac{\x_j^t - \x_k^t}{\|\x_j^t - \x_k^t\|}.
\label{eqn:sfm}
\end{align}
Here, $\v_j^*$ and $\tau_j$ are the desired velocity and relaxation time, respectively, whereas $\alpha_j$ and $\beta_j$ denote the strength and range of the interaction force. We have $\theta_j = \{\v_j^*, \tau_j, \alpha_j, \beta_j\}$.

Finally, we can assume a simple sensor observation model in which each sensor takes direct measurement of the target locations:
\begin{equation}
\z_{ij}^t = \x_j^t + \nu_{ij}^t, \quad \forall i \in \O_j^t.
\label{eqn:measurement}
\end{equation}
\end{example}

\section{Online Data-driven Distributed EKF}
\label{sec:method}

We now describe our technical approach to the proposed problem. Let $\T_i^t \triangleq \{j|i \in \O_j^t\}$ be the set of targets observed by sensor $i \in V$ at time $t$.
Using the augmented state $\y^t \triangleq [(\y_1^t)^T,\ldots,(\y_{m}^t)^T]^T$ and local output $\z_i^t=[(\z_{i1}^t)^T,\ldots,(\z_{i|\T_i^t|}^t)^T]^T$, system~\eqref{eq001} and output~\eqref{eq002} become
\begin{align}
\y^{t+1} = f(\y^t)+\eta^t, \qquad
\z_i^{t} = h_i(\y^t) + \nu_i^t \nonumber
\end{align}
where
\begin{align*}
&f(\y^t)\triangleq\left[
\begin{array}{c}
f_1(\y^t;\theta)\\
\vdots\\
f_m(\y^t;\theta)\\
\end{array}
\right],&&
\eta^t\triangleq\left[
\begin{array}{c}
\eta_1^t\\
\vdots\\
\eta_{m}^t\\
\end{array}
\right],\nonumber\\
&h_i(\y^t) \triangleq
\left[
\begin{array}{c}
h_{i1}(\y_1^t)\\
\vdots\\
h_{i|\T_i^t|}(\y_{|\T_i^t|}^t)\\
\end{array}
\right],
&&\nu_i^t \triangleq
\left[
\begin{array}{c}
\nu_{i1}^t\\
\vdots\\
\nu_{i|\T_i^t|}^t\\
\end{array}
\right].
\end{align*}

Recall that the functions $f$ and $h_i$ are unknown in Problem~\ref{prob1}.
To overcome the limited knowledge on the system, we regard functions $f$ and $h_i$ as regression functions and estimate them by utilizing Gaussian process regression (GPR). In GPR, knowledge on the regression function is expressed as data set, which is a set of input-output examples of the regression function.
At any time $t$, a set of input-output examples
$D_{f,i}^{t} = \langle I_{f,i}^{t},O_{f,i}^{t}\rangle$, $D_{h,i}^{t} = \langle I_{h,i}^{t},O_{h,i}^{t}\rangle$
are stored by sensor $i$, where $I_{f,i}^{t}$ is a set of inputs and $O_{f,i}^{t}$ is a set of outputs of function $f$ corresponding to the inputs. Likewise, $I_{h,i}^{t}$ is a set of inputs and $O_{h,i}^{t}$ is a set of outputs of function $h_i$ corresponding to the inputs.
We call $D_{f,i}^t$ and $D_{h,i}^t$ as data sets. They are initially empty, and are updated each time using estimates and outputs in an ``online'' fashion.

With the data-driven regression models, we can address Problem~\ref{prob1} in the Bayes filtering framework. We choose GP-EKF in this work and extend it to a distributed setting, but the same idea can be easily applied to other forms of GP-BayesFilters~\cite{ko2009gp}.
Algorithm~\ref{algo1} summarizes how the algorithm works in one episode. At any timestamp, each sensor node conducts state estimation given local information via GP-EKF (line 4), and then estimates are spread to neighboring nodes through a consensus algorithm (line 5).
Finally, the knowledge on the regression functions (\ie, datasets) is updated in lines 6-7.

In the following, we present the two key components of our method, namely the local GP-EKF and the consensus algorithm, in details.

\begin{algorithm}[t] \caption{Online Data-driven Distributed EKF}\label{algo1}
\begin{algorithmic}[1]

\STATE \textbf{Initialize}: target state $\hat{\y}^0_i$, information matrix $\Omega^0_i$, data set $D_{f,i}^{0}$$=\langle I_{f,i}^{0},O_{f,i}^{0}\rangle$, $D_{h,i}^{0}$$=\langle I_{h,i}^{0},O_{h,i}^{0}\rangle$;
\FOR{$t=1,2,\ldots,T/\Delta t$}
    \STATE \textbf{Input}: $\hat{\y}^{t-1}_i$, $\Omega^{t-1}_i$, $D_{f,i}^{t-1}$, $D_{h,i}^{t-1}$, measurement $\z^t_i$;
    \STATEx $\triangleright$ \textbf{\textit{Local GP-EKF}} (see Section~\ref{sec:method:ekf})
    \STATE Each sensor performs GP-EKF independently to obtain the prior
information $(q^{t|t-1}_i,\Omega^{t|t-1}_i)$ and the novel information $(\delta q^{t}_i,\delta \Omega^{t}_i)$;
    \STATEx $\triangleright$ \textbf{\textit{Hybrid Consensus}} (see Section~\ref{sec:method:consensus})
    \STATE Carry out consensus by iterating a number of regional averages on the two information pairs (prior and novel) and combine the fused pairs to obtain $(\hat{\y}_i^t$, $\Omega_i^t)$;
    \STATEx $\triangleright$ \textbf{\textit{Data set update}}
    \STATE $D_{f,i}^{t}=\langle [I_{f,i}^{t-1},\hat{\y}_i^{t-1}],[O_{f,i}^{t-1},\hat{\y}_i^{t}]\rangle$ for motion model;
    \STATE $D_{h,i}^{t}=\langle [I_{h,i}^{t-1},\hat{\y}^t],[O_{h,i}^{t-1},\z^t_i]\rangle$ for observation model; 
    \STATE \textbf{Return:} $\hat{\y}_i^t$, $\Omega_i^t$, $D_{f,i}^{t}$, $D_{h,i}^{t}$.
\ENDFOR
\end{algorithmic}
\end{algorithm}

\subsection{Local GP-EKF}
\label{sec:method:ekf}

GPR is an algorithm to estimate an unknown function under the assumption that the regression function is a Gaussian process (GP). Formally, stochastic process $f$ is Gaussian if $[f(\y_1),\cdots,f(\y_N)]^T$ is a multivariate Gaussian random variable, for any finite set of points $[\y_1,\cdots,\y_N]$~\cite{mackay2003information}.
GPR could return a mean function of outputs. In our method, the functions $f$ and $h_i$ are estimated with GPR mean function.

Take the motion function $f$ as an example. At time $t$, given the current dataset $D_{f,i}^{t-1}=\langle I_{f,i}^{t-1},O_{f,i}^{t-1}\rangle$ and previous state $\hat{\y}_i^{t-1}$, we want to predict the function output $f_*$. 
Under the assumption of GP, $O_{f,i}^{t-1}$ and $f_*$ follow a joint Gaussian distribution
\begin{align}
\mathcal{N}\left(\mathbf{0}, \begin{bmatrix}
K(I_{f,i}^{t-1},I_{f,i}^{t-1})+ \sigma_{\epsilon}^2 I & K(I_{f,i}^{t-1},\hat{\y}_i^{t-1})\\
K(\hat{\y}_i^{t-1},I_{f,i}^{t-1}) & K(\hat{\y}_i^{t-1},\hat{\y}_i^{t-1})
\end{bmatrix}\right) \nonumber
\end{align}
where $\sigma_{\epsilon}$ represents the noise level of output data. Function $K$ is kernel matrix (or covariance function) and the $(a,b)$ element of $K(I_{f,i}^{t-1},I_{f,i}^{t-1})$ is found by Gaussian kernel
\begin{align*}
K_{ab}&=k(I_{f,i}^{t-1}(a),I_{f,i}^{t-1}(b))\nonumber\\
&=\sigma_f^2 e^{-\frac{1}{2 l_f^2}(I_{f,i}^{t-1}(a)-I_{f,i}^{t-1}(b))^T(I_{f,i}^{t-1}(a)-I_{f,i}^{t-1}(b))}
\end{align*}
where $I_{f,i}^{t-1}(a)$ and $I_{f,i}^{t-1}(b)$ denote $a^{th}$, $b^{th}$ input data (column) of $I_{f,i}^{t-1}$ respectively.
Parameter $\sigma_f \in {\mathbb{R}}$ represents the scale of the outputs, and $l_f\in {\mathbb{R}}$ is the length scale.

\smallskip
\noindent\textbf{GPR mean function.} By the standard rules for conditioning Gaussians, the posterior $p(f_*|\hat{\y}_i^{t-1},D_{f,i}^{t-1})$ is
also a Gaussian distribution with mean function $\mu_f$
\begin{align}
\mu_f (\hat{\y}_i^{t-1},D_{f,i}^{t-1}) = O_{f,i}^{t-1}(K(I_{f,i}^{t-1},I_{f,i}^{t-1})+ \sigma_{\epsilon}^2 I)^{-1}k_*
\label{eq004}
\end{align}
where $k_*$$=$$K(I_{f,i}^{t-1},\hat{\y}_i^{t-1})$.
And Jacobian matrix of $\mu_f$ can be described by
\begin{align*}
\frac{\partial \mu_f (\hat{\y}_i^{t-1},D_{f,i}^{t-1})}{\partial \y}
&=\frac{\partial O_{f,i}^{t-1}(K(I_{f,i}^{t-1},I_{f,i}^{t-1})+ \sigma_{\epsilon}^2 I)^{-1}k_*}{\partial \y}\nonumber\\
&=O_{f,i}^{t-1}(K(I_{f,i}^{t-1},I_{f,i}^{t-1})+ \sigma_{\epsilon}^2 I)^{-1}\frac{\partial k_*}{\partial \y}.
\end{align*}
Similarly, one can also find GPR mean function $\mu_h (\hat{\y}_i^{t|t-1},D_{h,i}^{t-1})$ and Jacobian matrix $\frac{\partial \mu_h}{\partial \y}(\hat{\y}^{t|t-1}_i,D_{h,i}^{t-1})$ for observation function $h_i$.

\smallskip
\noindent\textbf{GPR covariance function.}
Covariance corresponding to the mean~\eqref{eq004} can be calculated by GPR covariance function:
\begin{align*}
    \Sigma_f(\hat{\y}_i^{t-1},D_{f,i}^{t-1}) = & k(\hat{\y}_i^{t-1},\hat{\y}_i^{t-1}) I \\
    &- k_*(K(I_{f,i}^{t-1},I_{f,i}^{t-1}) +\sigma_\epsilon^2 I)^{-1}(k_*)^T
\end{align*}
which represents covariance of the process $Q^t_i$.
The same method is used for computing $\Sigma_h(\hat{\y}^{t|t-1}_i,D_{h,i}^{t-1})$ to represent $R^t_i$.
Covariance matrices $Q^t_i$ and $R^t_i$ can be directly used in the Bayes filters.

As shown in Algorithm~\ref{algo2}, at every timestamp, we independently perform GP-EKF for each sensor. To facilitate information fusion, we adopt the information filter form and estimate the information matrix $\Omega^t_i=(P_i^t)^{-1}$  rather than covariance matrix $P_i^t$.
In particular, the pair $(q^{t|t-1}_i,\Omega^{t|t-1}_i)$ called prior information is found by dynamic propagation (lines 2-6). Then, the current measurement $\z^t_i$ is used to find the pair $(\delta q^{t}_i,\delta \Omega^{t}_i)$ called novel information (lines 7-11).

\begin{algorithm}[t] \caption{Local GP-EKF (Information filter form)}
\label{algo2}
\begin{algorithmic}[1]
    \STATE \textbf{Input}: $\hat{\y}^{t-1}_i$, $\Omega^{t-1}_i$, $D_{f,i}^{t-1}$, $D_{h,i}^{t-1}$, $\z^t_i$;
    \STATEx $\triangleright$ \textbf{\textit{Local Prediction}}
    \STATE GPR mean function: $\hat{\y}_i^{t|t-1} = \mu_f (\hat{\y}_i^{t-1},D_{f,i}^{t-1})$;
    \STATE Jacobian matrix: $A_i^t = \frac{\partial \mu_f}{\partial \y}(\hat{\y}_i^{t-1},D_{f,i}^{t-1})$;
    \STATE GPR covariance: $Q_i^t = \Sigma_f(\hat{\y}_i^{t-1},D_{f,i}^{t-1})$; $W_i^t=(Q_i^t)^{-1}$;
    \STATE  $\Omega_i^{t|t-1}$$=W_i^t-W_i^t A_i^t\left(\Omega_i^{t-1}+(A_i^t)^T W_i^t A_i^t\right)^{-1}(A_i^t)^T W_i^t$;
    or: $\Omega_i^{t|t-1}=\left(A_i^t(\Omega_i^{t-1})^{-1}(A_i^t)^T+Q_i^t\right)^{-1}$;
    \STATE Information vector: $q_i^{t|t-1} = \Omega_i^{t|t-1}\hat{\y}_i^{t|t-1}$;
    
    \STATEx $\triangleright$ \textbf{\textit{Local Correction}}
    \STATE Jacobian matrix: $C^t_i=\frac{\partial \mu_{h}}{\partial \y}(\hat{\y}_i^{t|t-1},D_{h,i}^{t-1})$;
    \STATE GPR covariance: $R^t_i = \Sigma_{h}(\hat{\y}_i^{t|t-1},D_{h,i}^{t-1})$;
    \STATE $\bar{\z}^t_i = \z^t_i -\mu_h(\hat{\y}_i^{t|t-1},D_{h,i}^{t-1})+C^t_i\hat{\y}_i^{t|t-1}$;
    \STATE $\delta q^t_i = (C^t_i)^T (R^t_i)^{-1} \bar{\z}^t_i$;
    \STATE $\delta \Omega^t_i = (C^t_i)^T (R^t_i)^{-1} C^t_i$;
    \STATE \textbf{Output:} $q_i^{t|t-1}$, $\Omega_i^{t|t-1}$, $\delta q^t_i$, $\delta \Omega^t_i$.
\end{algorithmic}
\end{algorithm}

\subsection{Hybrid Consensus}
\label{sec:method:consensus}

As we discussed earlier, in practice, the observations of an individual sensor may be incomplete due to its perception range and occlusions, and also subject to noises and errors. Therefore, it is necessary to fuse information from different sensors for accurate state estimation.

In the literature, consensus is a widely used technique for distributed computations over networks. In this work, we adopt a special form of consensus with stability guarantee called hybrid consensus~\cite{BattistelliC16}. As shown in Algorithm~\ref{algo3}, each node iteratively recalculates prior information $(q^{t|t-1}_i,\Omega^{t|t-1}_i)$ and novel information $(\delta q^{t}_i,\delta \Omega^{t}_i)$ through communicating with its neighbors $N_i$, where $L \in {\mathbb{N}}$ is the number of iterations (lines 4-11).
A set of weights $\pi_{ii'}\geq 0$ for $i' \in N_i$ is used for the recalculation, and is a convex combination, \ie, $\sum_{i' \in N_i\bigcup \{i\}} \pi_{ii'}=1$. Without loss of generality, we choose uniform weights in this work: $\pi_{ij} = 1/(|N_i |+1)$ for $j\in N_i\bigcup \{i\}$.
Then, the filtered estimate $\hat{\y}_i^t$ is computed by the consentaneous information (lines 12-14).


\begin{algorithm}[t] \caption{Hybrid Consensus}\label{algo3}
\begin{algorithmic}[1]
    \STATE \textbf{Input}: $q_i^{t|t-1}$, $\Omega_i^{t|t-1}$, $\delta q^t_i$, $\delta \Omega^t_i$;
    \STATEx $\triangleright$ \textbf{\textit{Consensus}}
    \STATE Prior info.: $q^{t|t-1}_i(0)$$=q^{t|t-1}_i$;  $\Omega^{t|t-1}_i(0)$$=\Omega^{t|t-1}_i$;
    \STATE Novel info.: $\delta q^t_i(0)$$=\delta q^t_i$;  $\delta \Omega^t_i(0)$$=\delta \Omega^t_i$;
    \FOR{$\ell=0,\ldots,L-1$}
        \STATE Send  $q^{t|t-1}_i(\ell)$, $\Omega^{t|t-1}_i(\ell)$, $\delta q^{t}_i(\ell)$, $\delta\Omega^{t}_i(\ell)$ to $i' \in N_i$;
        \STATE \scalebox{.95}[1.0]{Get $q^{t|t-1}_{i'}(\ell)$, $\Omega^{t|t-1}_{i'}(\ell)$, $\delta q^{t}_{i'}(\ell)$, $\delta\Omega^{t}_{i'}(\ell)$ from $i' \in N_i$;}
        \STATE $q^{t|t-1}_i(\ell+1) = \sum_{i' \in N_i\bigcup\{i\}}\pi^{ii'}q^{t|t-1}_{i'}(\ell)$;
        \STATE $\Omega^{t|t-1}_i(\ell+1)= \sum_{i' \in N_i\bigcup\{i\}}\pi^{ii'}\Omega^{t|t-1}_{i'}(\ell)$;
        \STATE $\delta q^t_i(\ell+1) = \sum_{i' \in N_i\bigcup\{i\}}\pi^{ii'}\delta q^{t}_{i'}(\ell)$;
        \STATE $\delta \Omega^t_i(\ell+1)= \sum_{i' \in N_i\bigcup\{i\}}\pi^{ii'}\delta\Omega^t_{i'}(\ell)$;
    \ENDFOR
    \STATEx $\triangleright$ \textbf{\textit{Estimation}}
    \STATE $q^t_i = q^{t|t-1}_i(L) +  \delta q^t_i(L)$;
    \STATE $\Omega^t_i = \Omega^{t|t-1}_i(L) + \delta\Omega^t_i(L)$;
    \STATE $\hat{\y}^t_i$$=(\Omega^t_i)^{-1}q^t_i$;
    \STATE \textbf{Output:} $\hat{\y}_i^t$, $\Omega_i^t$.
\end{algorithmic}
\end{algorithm}



\section{Experiments on Synthetic Data}

In this section, we evaluate our method on synthetic data generated by simulating the movement and interaction of traffic agents (\eg, pedestrians). Assume that the agents are tracked by a sensor network of four \emph{localization sensors} and six \emph{communication nodes}, which are randomly distributed within an $8.0\times 8.0$ square area, as depicted in Fig.~\ref{fig:two_obj_scene}. Note that both types of nodes can process local information and exchange information with their linked neighbors, but only the localization sensors can make measurements about the targets. The effective range of a sensor is $4.0$. 

\subsection{Data Generation}
\label{sec:syn:data}

Given $m$ targets, we simulate the trajectory of each target using the social force model with parameters $\{\v_j^*, \tau, \alpha, \beta\}$, as described in Example~\ref{sec:pro:example}. Each target has its own preferred velocity, but shares the remaining parameters with other targets. The default values are $\tau=0.25, \alpha=6, \beta= 5$. 

\begin{figure}[t]
\centering
\includegraphics[width=1.0\linewidth]{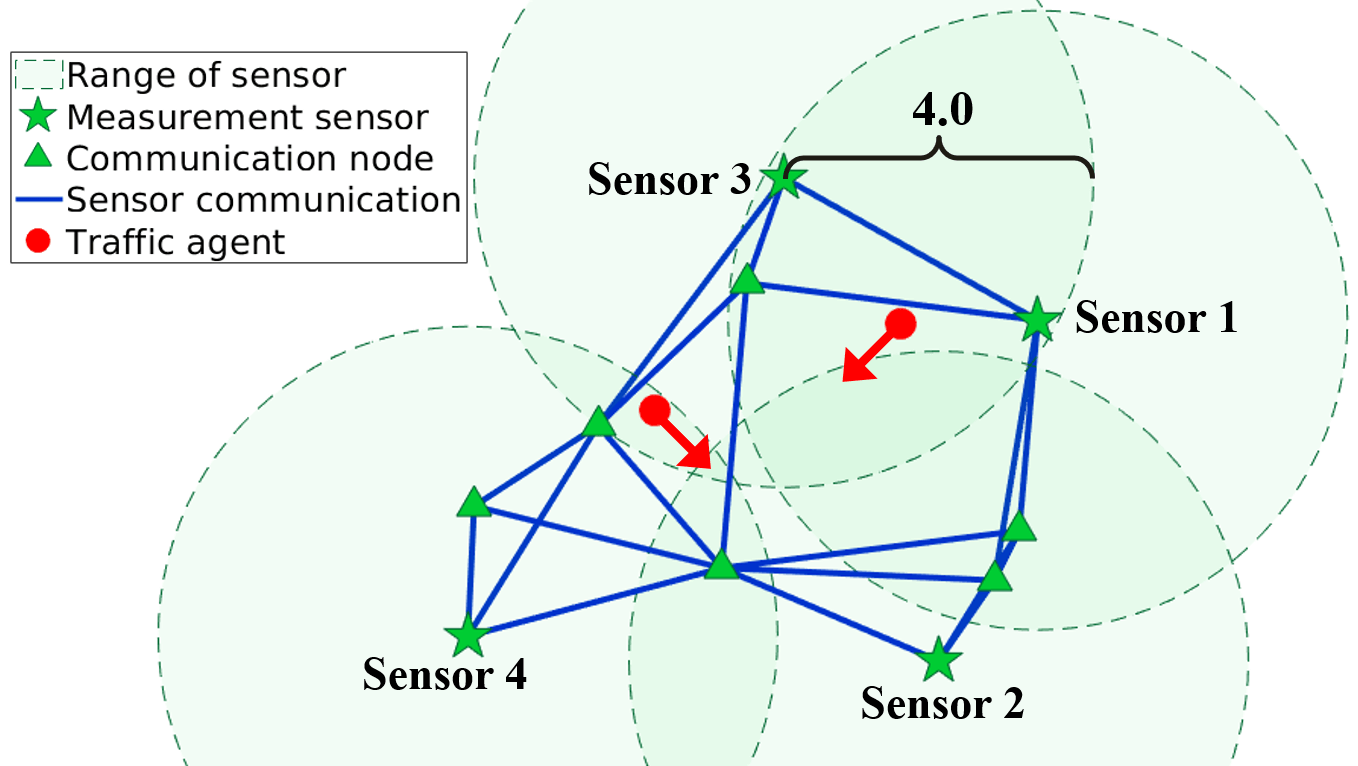}
\caption{Synthetic scenario of interaction between two traffic agents. The agents are tracked by a sensor network comprising four measurement sensors and six communication nodes.}
\label{fig:two_obj_scene}
\end{figure}

We consider a typical case of $m=2$. As shown in Fig.~\ref{fig:two_obj_scene}, the two targets start from different random positions in two $1.5\times 1.5$ square areas and move to the center area simultaneously. The desired velocities are $\v_1^*=(1,-1)$ and $\v_2^*=(-1,-1)$, respectively. We add small noise $\eta \sim \N(0, \sigma_{\eta}^2)$ with $\sigma_{\eta}=0.001$ in Eq.~\eqref{eq:motion_eq}.

To conduct simultaneous state estimation and behavior learning, we run the simulation to generate a large number of episodes of trajectories. The length of each episode is $T$ with time step $\Delta t$, including $(T/ \Delta t)+1$ states $\y^t$ ($t\in\{0,1,\dots,T/ \Delta t\}$). We set $T=3$ and $\Delta t = 0.25$ in this experiment. Finally, the measurement model is given in Eq.~\eqref{eqn:measurement} with noise $\nu \sim \N(0,\sigma_{\nu}^2)$ where $\sigma_{\nu} = 0.2$.

\subsection{Motion Models}
\label{sec:syn:models}
We compare the following motion models in EKF:
\begin{itemize}

\item {\bf SFM}. The social force model as described in Eq.~\eqref{eq:motion_eq} and Eq.~\eqref{eqn:sfm}, \ie, the true motion model. All parameters are assumed to be known, as described in Section~\ref{sec:syn:data}. 

\item {\bf LIN}. A linear constant velocity model which predicts the state of each target independently
\begin{align}
\x_j^{t+1} = \x_j^t + \v_j^t \cdot \Delta t, \qquad \v_j^{t+1} = \v_j^t. \nonumber
\end{align}


We assume the velocity of each target $\v_j^*$ is given. To accommodate the deviation from the true motion model, we use a higher process noise $\sigma_{\eta}=0.1$ with LIN.

\item {\bf GP}. The Gaussian process regression model introduced in Section~\ref{sec:method}. The model hyperparameters are $\{\sigma_f, l_f, \sigma_{\epsilon}\}$, which are often estimated by maximum marginal likelihood~\cite{chen2018priors}. To accommodate online learning, we manually tune the hyperparameters and set $\sigma_f=1$, $l_f=2$, and $\sigma_{\epsilon}=0.5$ by default.

\end{itemize}


\subsection{Experimental Protocol and Evaluation Metrics}
\label{sec:syn:protocol}

In this experiment, we generate $M$ episodes of trajectories for training, and $N$ additional episodes for testing. In each episode, we initialize the state and the corresponding information matrix in Algorithm~\ref{algo1} with the first two observations as follows:
\begin{align}
\hat{\y}_i^0
=\begin{pmatrix} \z_i^0 \\ \frac{\z_i^1-\z_i^0}{\Delta t}\end{pmatrix}, \;
\Omega_i^0 =
\begin{bmatrix}
 \sigma_{\nu}^2 &  &  & \\ 
 & \sigma_{\nu}^2 &  & \\ 
 &  & \frac{2\sigma_{\nu}^2}{(\Delta t)^2} & \\ 
 &  &  & \frac{2\sigma_{\nu}^2}{(\Delta t)^2}
\end{bmatrix}^{-1}. \nonumber
\label{eqn:init}
\end{align}
For GP, in the first training episode, we also compute the second state $\hat{\y}_i^1$ to initialize the data set: $D_{f,i}^{1}$$= \langle \hat{\y}_i^0,\hat{\y}_i^1\rangle$.  

We use the root mean squared error (RMSE) of position to quantify the performance of each method. For each episode, the RMSE of the trajectory is computed as
\begin{align}
RMSE=\frac{\Delta t}{mT}\sum\limits_{t=1}^{T/\Delta t}\sum\limits_{i=1}^{m}\|\x^t_i-\hat{\x}^t_i\|. \nonumber
\end{align}
The averaged RMSE of all testing episodes with standard deviation is computed as the overall performance metrics.

\subsection{Experiment Results}

We have generated 250 episodes of trajectories and randomly split them into 200 training and 50 testing episodes. The default value of $L$ is set to 4.
We first compare the results obtained by running EKF on each sensor individually with those obtained by the consensus algorithm. 
Table~\ref{table:gt} reports the average RMSE of different methods on the testing episodes. The results show that, due to the limited sensing range, running EKF on individual sensors yields large estimation errors. And GP performs worse than LIN, since GP relies on the observations to learn the motion model. With consensus, the performance improves significantly for all motion models. Further, GP achieves lower error than LIN, which suggests the importance of consensus in learning the motion model. We also note that, with consensus, all sensors have almost identical errors, indicating that the consensus algorithm converges with $L=4$ for this sensor network topology.


\setlength{\textfloatsep}{0.3cm}
\setlength{\tabcolsep}{4pt}
\begin{table}[tb]
\small
\centering
\caption{Tracking RMSE of distributed EKF on synthetic data.}
\label{table:gt}
\begin{tabular}{|c|c|c|c|c|}
\hline
& & GP & LIN & SFM\\
\hline
\parbox[t]{3mm}{\multirow{5}{*}{\rotatebox[origin=c]{90}{Individual}}} & Sensor 1 & $\mathbf{1.24}\pm 0.23$ & $\mathbf{1.29}\pm 0.72$ & $\mathbf{0.37}\pm 0.23$\\
& Sensor 2 & $\mathbf{0.81}\pm 0.40$ & $\mathbf{0.66}\pm 0.51$ & $\mathbf{0.26}\pm 0.14$\\
& Sensor 3 & $\mathbf{0.68}\pm 0.18$ & $\mathbf{0.47}\pm 0.22$ & $\mathbf{0.16}\pm 0.06$\\
& Sensor 4 & $\mathbf{1.80}\pm 0.43$ & $\mathbf{1.02}\pm 0.51$ & $\mathbf{0.26}\pm 0.06$\\
\cline{2-5}
& Average & $\mathbf{1.13}\pm 0.31$ & $\mathbf{0.86}\pm 0.49$ & $\mathbf{0.26}\pm 0.12$\\
\hline
\parbox[t]{3mm}{\multirow{5}{*}{\rotatebox[origin=c]{90}{Consensus}}} & Sensor 1 & $\mathbf{0.14}\pm 0.04$ & $\mathbf{0.16}\pm 0.03$ & $\mathbf{0.14}\pm 0.10$\\
& Sensor 2 & $\mathbf{0.14}\pm 0.04$ & $\mathbf{0.16}\pm 0.03$ & $\mathbf{0.14}\pm 0.10$\\
& Sensor 3 & $\mathbf{0.14}\pm 0.04$ & $\mathbf{0.16}\pm 0.03$ & $\mathbf{0.14}\pm 0.10$\\
& Sensor 4 & $\mathbf{0.14}\pm 0.04$ & $\mathbf{0.16}\pm 0.03$ & $\mathbf{0.14}\pm 0.10$\\
\cline{2-5}
& Average & $\mathbf{0.14}\pm 0.04$ & $\mathbf{0.16}\pm 0.03$ & $\mathbf{0.14}\pm 0.10$\\
\hline
\end{tabular}
\end{table}

\begin{figure}[tb]
\centering
\includegraphics[width=0.9\linewidth]{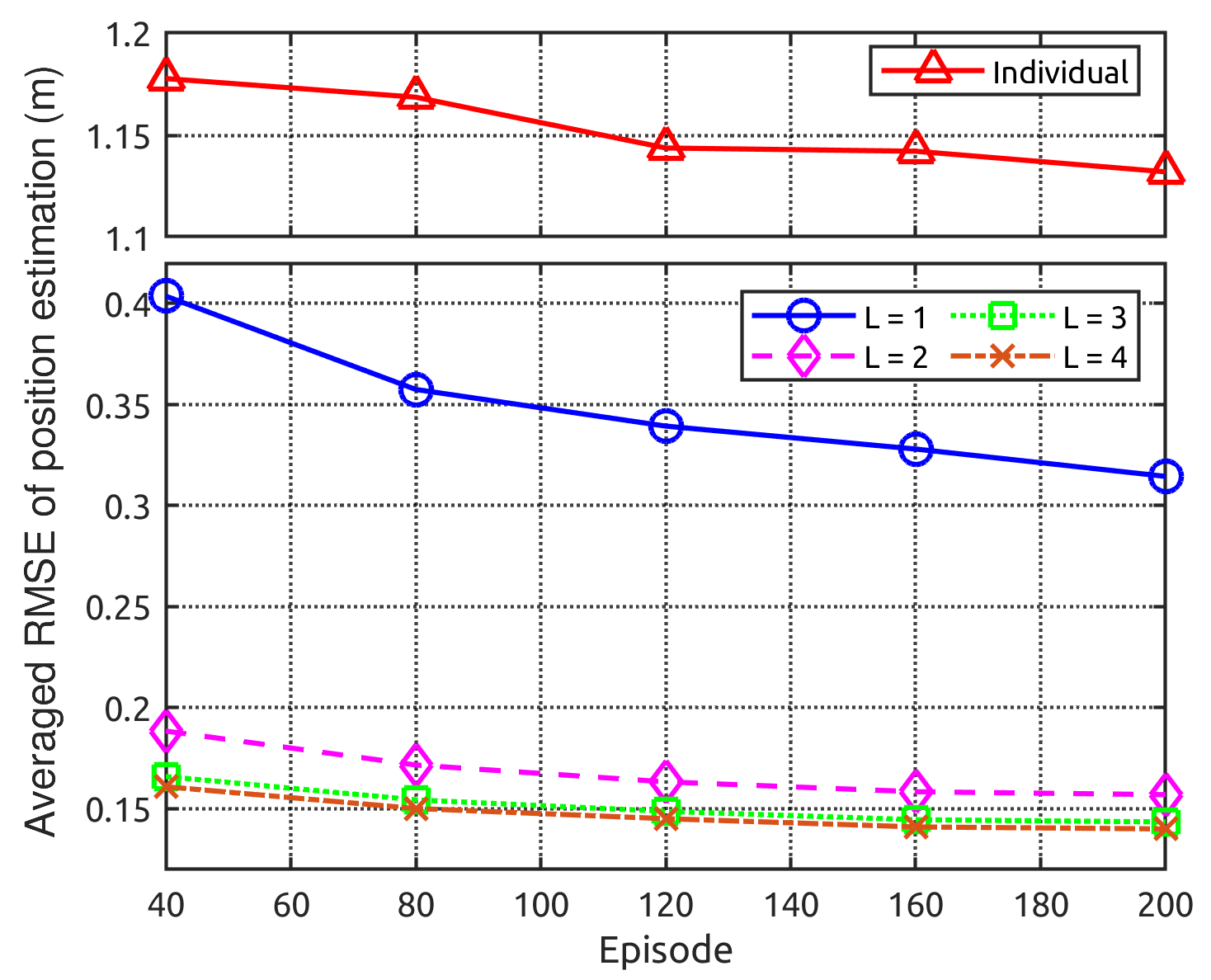}
\caption{Tracking performance of GP-EKF w.r.t. the number of training episodes and number of consensus iterations $L$.}
\label{fig:rmse_trend}
\end{figure}

Figure~\ref{fig:rmse_trend} plots the RMSE curves of GP-EKF w.r.t. two important parameters, namely the number of training episodes and $L$. As one can see, the error decreases as the number of training episodes increases, which indicates that GP is able to learn the movement behavior online. Further, comparing the RMSE curves for different choices of $L$, we can see that the consensus algorithm converges when $L\geq 3$.

We also conduct experiment to test the quality of the learned motion models, which are the critical components for Bayes filters. In this experiment, we provide each model with the ground truth state at $t=0$, ask it to predict the states for $t = 1, \ldots, T/\Delta t$, and compute the average RMSE over all test sequences with standard deviation. The results are summarized in Table~\ref{table:prediction}. As can be seen, the GP (con.) models learned through running EKF with consensus are significantly better than the GP (ind.) models, which are learned by each sensor individually. And the GP (con.) models also significantly outperform the LIN models, verifying the effectiveness of our data-driven approach.

\begin{table}[t]
\small
\centering
\caption{Prediction RMSE on synthetic data.}
\label{table:prediction}
\begin{tabular}{|c|c|c|c|c|}
\hline
& GP (ind.) & GP (con.) & LIN & SFM\\
\hline
S1 & $\mathbf{1.38}\pm 0.36$ & $\mathbf{0.24}\pm 0.10$ & $\mathbf{1.30}\pm 0.11$ & $\mathbf{0.004}\pm 0.002$\\
S2 & $\mathbf{1.29}\pm 0.37$ & $\mathbf{0.24}\pm 0.10$ & $\mathbf{1.30}\pm 0.11$ & $\mathbf{0.004}\pm 0.002$\\
S3 & $\mathbf{0.73}\pm 0.19$ & $\mathbf{0.24}\pm 0.10$ & $\mathbf{1.30}\pm 0.11$ & $\mathbf{0.004}\pm 0.002$\\
S4 & $\mathbf{2.11}\pm 0.43$ & $\mathbf{0.24}\pm 0.10$ & $\mathbf{1.30}\pm 0.11$ & $\mathbf{0.004}\pm 0.002$\\
\hline
Avg & $\mathbf{1.38}\pm 0.34$ & $\mathbf{0.24}\pm 0.10$ & $\mathbf{1.30}\pm 0.11$ & $\mathbf{0.004}\pm 0.002$\\
\hline
\end{tabular}
\end{table}

\section{Experiments on Robot Platform}

To assess the effectiveness of the proposed method on real system, we have also conducted experiments on a robot platform with three Khepera III robots \cite{kheperaIII2013}.

\subsection{Experiment Settings}

\begin{figure}[t]
\centering
\includegraphics[width=0.9\linewidth]{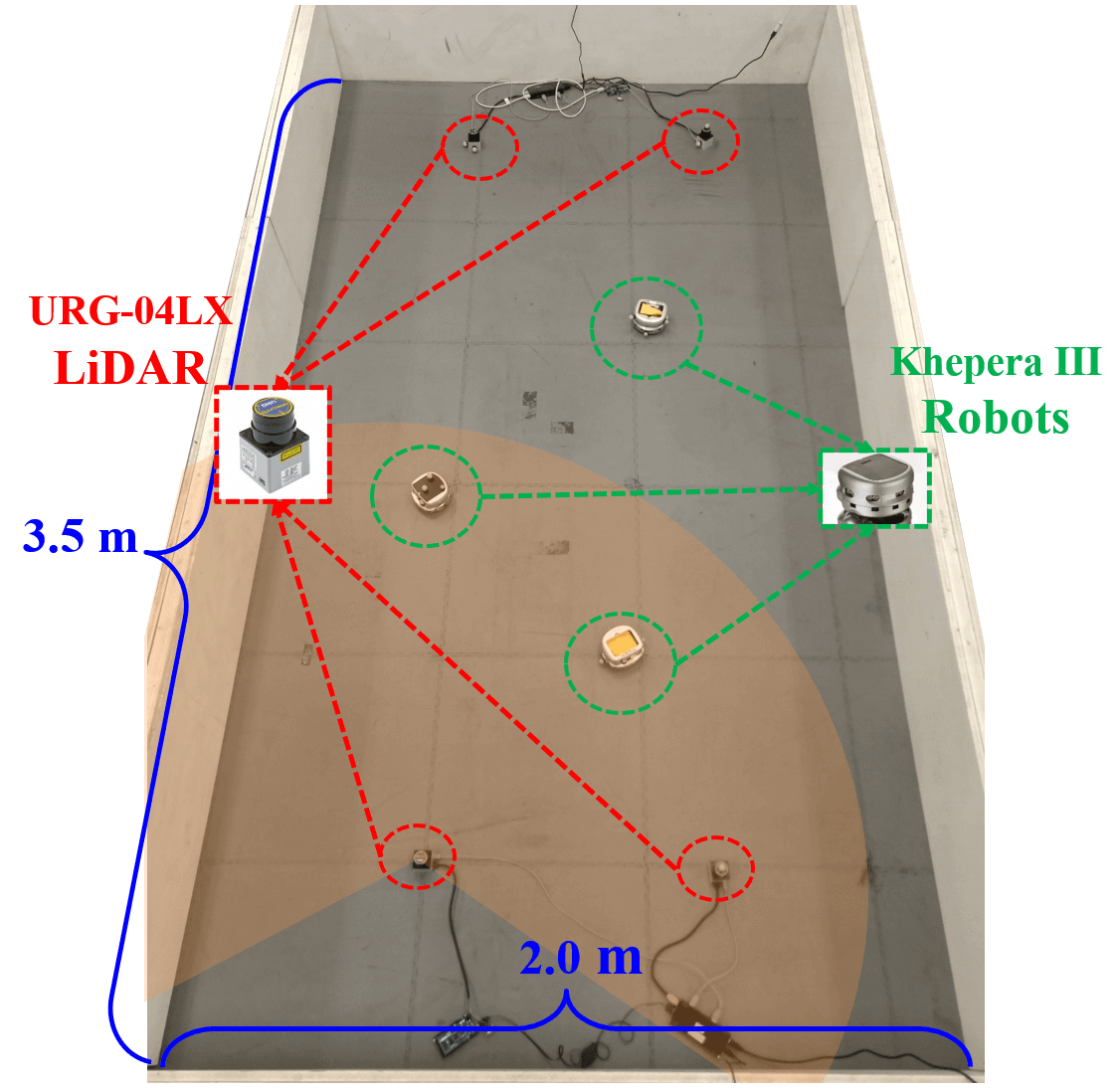}
\caption{The Khepera III robot platform.}
\label{fig:robot}
\end{figure}

As shown in Fig.~\ref{fig:robot}, the arena is $3.5$ meters long and $2.0$ meters wide. The Khepera III robot has the size of 115mm$\times$155mm$\times$108mm and constructs differential drive dynamics. Each robot is programmed to move according to the social force model as described in Example~\ref{sec:pro:example}. The model parameters are chosen as follows: $\tau=1.0, \alpha=0.24, \beta=0.5$. The desired velocities of the robots are $\v_1^*=(0.04,0.04)$m/s, $\v_2^*=(0.08,0.08)$m/s, and $\v_3^*=(0.08,-0.08)$m/s. The scenario is designed to simulate the interactions among agents when crossing an intersection. 

The locations of the robots are tracked by a VICON motion capture system with high accuracy. The system runs up to 1000 Hz, so we could also estimate the velocities based on the high-frequency location records. The locations and velocities obtained from VICON are considered as the ground truth states. As shown in Fig.~\ref{fig:robot_predict}, the robot locations obtained from VICON deviate from those predicted based on the ideal SFM model. The reason is that, in practice, the robot motion trajectory can be influenced by many un-modeled factors, such as frictional resistance, limitation of two-wheel motion, and so on.

Note that the ground truth states are used for evaluation purpose only. To conduct this experiment, we further obtain robot location measurement from four Hokuyo URG-04LX 2D LiDARs \cite{hokuyo2005} placed in the arena. The Hokuyo LiDARs scan a 2D plane with an angle of 240 degrees and 0.36 angular resolution. The scan rate is 10 scans/sec and maximum range is 4.0 m. We design the arena to be enclosed by four walls to facilitate the LiDAR sensing. For each episode, we take 12 measurements with a time interval of 1.2s. The raw LiDAR measurements are first transferred into $(x,y)$-coordinates based on the position and orientation of each LiDAR. Then we rule out all points near the walls, apply DBSCAN clustering~\cite{EsterKSX96} to the remaining points, obtain the location measurements by computing the centers of the three largest clusters, and finally associate the measurements with the three robots according to their previous locations using the Hungarian algorithm. Based on the collected data, the measurement noise is set to $\sigma_{\nu} = 0.2$. In this experiment, we suppose each LiDAR can communicate with other two nearest LiDARs.

\subsection{Experiment Results}
For this experiment, we have collected 153 episodes and randomly split them into 128 training and 25 testing episodes. For all motion models, we adopt the same parameters as in Section~\ref{sec:syn:models}. These models are evaluated on the robot data using the same protocol as in Section~\ref{sec:syn:protocol}. Table~\ref{table:tracking-robot} reports the tracking RMSE of different models in the distributed EKF framework, and Table~\ref{table:predict-robot} reports the prediction results. As can be seen, the GP model not only outperforms the LIN model, but is also better than the ``true'' SFM model. As we mentioned above, real-world robot trajectories are influenced by other hard-to-model factors (\eg, frictional resistance). The idealized SFM model ignores these factors while the GP model is able to learn them from observed real-world data. It explains why the performance of GP is on par with that of SFM on synthetic data but better than SFM on robotic data. It also demonstrates the reason why data-driven methods outperform idealized models on real-world experiments.


\begin{table}[t]
\small
\centering
\caption{Tracking RMSE of distributed EKF on Robot data.}
\label{table:tracking-robot}
\begin{tabular}{|c|c|c|c|}
\hline
& GP & LIN & SFM\\
\hline
Sensor 1 & $\mathbf{0.202}\pm 0.092$ & $\mathbf{0.203}\pm 0.103$ & $\mathbf{0.236}\pm 0.043$\\
Sensor 2 & $\mathbf{0.204}\pm 0.094$ & $\mathbf{0.206}\pm 0.106$ & $\mathbf{0.236}\pm 0.043$\\
Sensor 3 & $\mathbf{0.205}\pm 0.094$ & $\mathbf{0.208}\pm 0.107$ & $\mathbf{0.236}\pm 0.043$\\
Sensor 4 & $\mathbf{0.202}\pm 0.089$ & $\mathbf{0.202}\pm 0.098$ & $\mathbf{0.236}\pm 0.042$\\
\hline
Average & $\mathbf{0.203}\pm 0.092$ & $\mathbf{0.205}\pm 0.103$ & $\mathbf{0.236}\pm 0.043$\\
\hline
\end{tabular}
\end{table}

\begin{table}[t]
\small
\centering
\caption{Prediction RMSE on Robot data.}
\label{table:predict-robot}
\begin{tabular}{|c|c|c|c|}
\hline
& GP & LIN & SFM\\
\hline
Sensor 1 & $\mathbf{0.125}\pm 0.024$ & $\mathbf{0.688}\pm 0.045$ & $\mathbf{0.242}\pm 0.044$\\
Sensor 2 & $\mathbf{0.127}\pm 0.024$ & $\mathbf{0.688}\pm 0.045$ & $\mathbf{0.242}\pm 0.044$\\
Sensor 3 & $\mathbf{0.127}\pm 0.024$ & $\mathbf{0.688}\pm 0.045$ & $\mathbf{0.242}\pm 0.044$\\
Sensor 4 & $\mathbf{0.126}\pm 0.024$ & $\mathbf{0.688}\pm 0.045$ & $\mathbf{0.242}\pm 0.044$\\
\hline
Average & $\mathbf{0.126}\pm 0.024$ & $\mathbf{0.688}\pm 0.045$ & $\mathbf{0.242}\pm 0.044$\\
\hline
\end{tabular}
\end{table}

\nop{
\begin{table}[t]
\small
\centering
\caption{Average tracking and prediction RMSE on robot data.}
\label{table:result-robot}
\begin{tabular}{|c|c|c|c|}
\hline
& GP & LIN & SFM\\
\hline
Tracking & $\mathbf{0.20}\pm 0.09$ & $\mathbf{0.21}\pm 0.10$ & $\mathbf{0.20}\pm 0.09$\\
\hline
Prediction & $\mathbf{0.13}\pm 0.02$ & $\mathbf{0.69}\pm 0.04$ & $\mathbf{0.24}\pm 0.04$ \\
\hline
\end{tabular}
\end{table}
}

\nop{
\setlength{\textfloatsep}{0.3cm}
\setlength{\tabcolsep}{4pt}
\begin{table}[tb]
\small
\centering
\caption{Tracking RMSE of distributed EKF on robotic data.}
\label{table:gt-robot}
\begin{tabular}{|c|c|c|c|}
\hline
& GP & LIN & SFM\\
\hline
Sensor 1 & 0.0702 & 0.0764 & 0.0741\\
Sensor 2 & 0.0699 & 0.0756 & 0.0736\\
Sensor 3 & 0.0691 & 0.0746 & 0.0740\\
Sensor 4 & 0.0703 & 0.0764 & 0.0739\\
\hline
Average & 0.0699 & 0.0757 & 0.0739\\
\hline
\end{tabular}
\end{table}
}


\begin{figure}[t]
\centering{
\subfigure[Trajectory 1]
{
   \label{fig:robot-traj1}
   \includegraphics[width=0.45\linewidth]{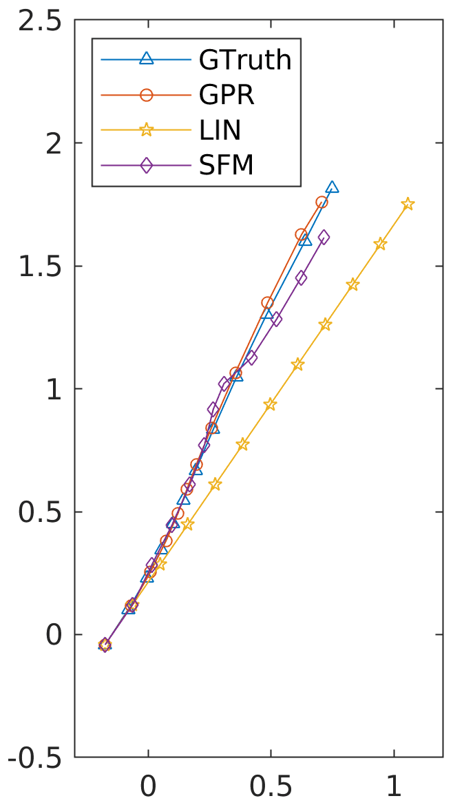}
}
\subfigure[Trajectory 2]
{
   \label{fig:robot-traj2}
   \includegraphics[width=0.45\linewidth]{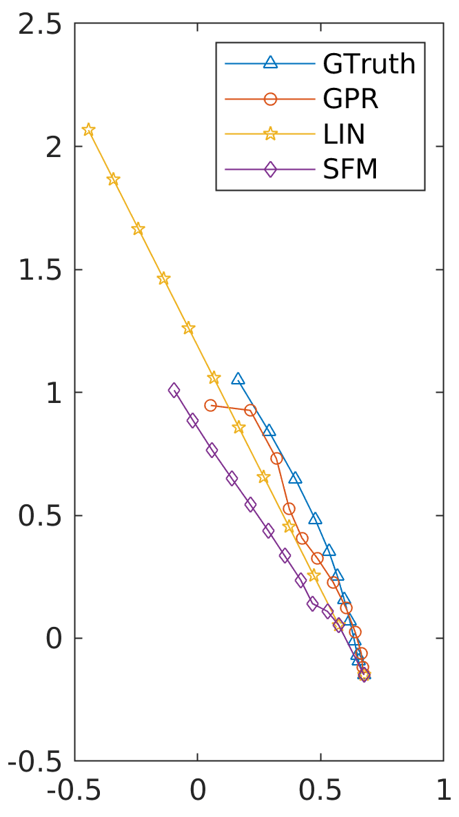}
}%
}
\caption{Comparison of the predicted trajectories of different motion models.}
\label{fig:robot_predict}
\end{figure}

In Fig.~\ref{fig:robot_predict}, we further provide two representative examples of the trajectories predicted by different motion models. For better clarity, we only visualize the trajectories of one robot in each plot. In the designed scenario, every robot is trying to maintain the preferred velocity while avoiding collision with others.
A common resulting phenomenon is that a robot would slow down or adjust the orientation if there is another robot in front of it.
It is not surprising that LIN is neither able to predict the shift in direction (Fig.~\ref{fig:robot-traj1}) nor the change in speed (Fig.~\ref{fig:robot-traj2}). The SFM is able to model the ``ideal'' interactions among agents but ignores the aforementioned hard-to-model factors in real-world experiments. In contrast, GPR directly learns the multi-agent movement patterns from the noisy observations. As seen in Fig.~\ref{fig:robot_predict}, GPR produces the best predictions among all motion models. The experiment results thus show the advantage of online GP models in learning real-world motion patterns.

\subsection{Comparison with State-of-the-Art Offline Prediction Model}
Recently, deep neural networks, in particular long short-term memory (LSTM) models~\cite{AlahiGRRLS16}, are shown to achieve the state-of-the-art performance in predicting complex pedestrian movements. 
But unlike our problem setting in which the motion model (\ie, GP) is learned online from noisy EKF estimations, existing work (\eg, \cite{AlahiGRRLS16}) learns LSTM offline with human-annotated noise-free movement data.

In this experiment, we train the LSTM model on the robot data and compare it with our GP model in terms of their ability to \emph{learn from small-scale noisy data}. Specifically, to prepare data to train LSTM offline, we use the same 128 training episodes and fuse the four LiDAR measurements by using consensus on measurement (CM) method~\cite{olfati2005distributed}. Following~\cite{AlahiGRRLS16}, we adopt an encoder-decoder sequence-to-sequence network architecture. The observed 2D points are first fed into a multi-layer perceptron (MLP) to obtain 64-dim feature embeddings. The encoder LSTM transforms the embedding into a 64-dim hidden state as the input of the decoder. The decoder LSTM transforms the 64-dim feature into a new 64-dim embedding. Then, another MLP converts decoder embedding into 2D vectors as the output prediction. The whole network is trained end-to-end using Adam~\cite{KingmaB14} optimizer for 50 epochs with a learning rate of $1\times 10^{-3}$ and batch size of 8. The $L2$ loss is adopted to minimize the distance between the future measurements and predicted locations. The lengths of the input and output sequences are $2$ and $10$, respectively.

\begin{figure}[t]
\centering
\includegraphics[width=1.0\linewidth]{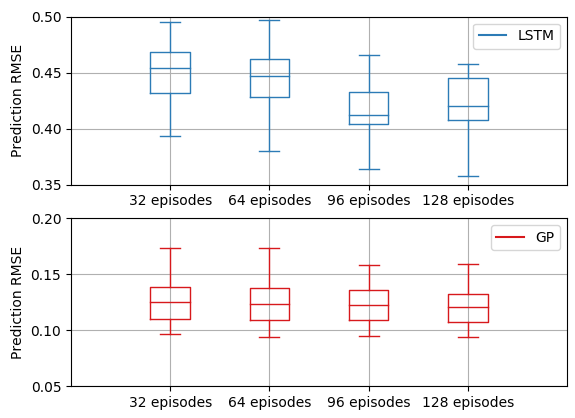}
\caption{Comparison of prediction RMSEs of LSTM and GP with different numbers of training episodes.}
\label{fig:lstm-gp}
\end{figure}

During testing, we follow the previous protocol to test the learnt LSTM model on same 25 testing episodes. In each episode, we provide LSTM with the ground truth state at $t=0$. Based on the constant velocity assumption, we can estimate the location for $t=1$ and utilize the trained sequence-to-sequence LSTM to predict the next $10$ locations.
Figure~\ref{fig:lstm-gp} shows the prediction results of the GP and LSTM models trained with varying number of episodes. As can be seen, GP significantly outperforms LSTM in this experiment. Besides, GP achieves more stable prediction performance even with a small number of training episodes (\eg, 32 episodes), which demonstrates the advantage of GP in learning from small-scale noisy data.

\addtolength{\textheight}{-4cm}   


\section{Conclusion and Future Work}

In this paper, we have studied the problem of jointly estimating the states of dynamic objects and learning their movement behaviors in a sensor network. A simple yet effective solution is introduced, in which we extend the family of GP-BayesFilters to a distributed setting, and show that the GP models can be learned online with noisy sensor observations. Experiment results on both synthetic data and data simulated using a multi-robot platform to verify the effectiveness of our method.

While our problem formulation is quite general, assuming no knowledge about the motion and observation models, it still can be extended in multiple ways according to potential real-world scenarios. \emph{First}, other noise models may be considered. For example, the observations may contain outliers, caused by sensor failures. \emph{Second}, data association in multi-target tracking~\cite{5338565}, \ie, associating all observations which belong to the same target, must be addressed. And the problem becomes more challenging for a sensor network, as the association across views needs to be further computed~\cite{KamalBFR16}. \emph{Third}, dynamic objects may change their movement behaviors over time, \eg, a pedestrian stops to greet a friend on the sidewalk. We leave these directions to future research.

As to the technical approach, we plan to investigate fast methods for GP-based filters (\eg,~\cite{PanYTB17}) in order to maintain affordable computation with large-scale datasets. Finally, for the consensus filters, it is interesting to further study the cases where the communication topology and thus the weights are time-varying~\cite{ZhuM10}.


\bibliographystyle{IEEEtran}
\bibliography{IEEEabrv,distributed-gpr-ekf}

\begin{thebibliography}{10}
\providecommand{\url}[1]{#1}
\csname url@rmstyle\endcsname
\providecommand{\newblock}{\relax}
\providecommand{\bibinfo}[2]{#2}
\providecommand\BIBentrySTDinterwordspacing{\spaceskip=0pt\relax}
\providecommand\BIBentryALTinterwordstretchfactor{4}
\providecommand\BIBentryALTinterwordspacing{\spaceskip=\fontdimen2\font plus
\BIBentryALTinterwordstretchfactor\fontdimen3\font minus
  \fontdimen4\font\relax}
\providecommand\BIBforeignlanguage[2]{{%
\expandafter\ifx\csname l@#1\endcsname\relax
\typeout{** WARNING: IEEEtran.bst: No hyphenation pattern has been}%
\typeout{** loaded for the language `#1'. Using the pattern for}%
\typeout{** the default language instead.}%
\else
\language=\csname l@#1\endcsname
\fi
#2}}

\bibitem{ko2009gp}
J.~Ko and D.~Fox, ``{GP}-{B}ayes{F}ilters: Bayesian filtering using {Gaussian}
  process prediction and observation models,'' \emph{Autonomous Robots},
  vol.~27, no.~1, pp. 75--90, 2009.

\bibitem{RasmussenW06}
C.~E. Rasmussen and C.~K.~I. Williams, \emph{{Gaussian} processes for machine
  learning}.\hskip 1em plus 0.5em minus 0.4em\relax {MIT} Press, 2006.

\bibitem{Olfati-SaberFM07}
R.~Olfati{-}Saber, J.~A. Fax, and R.~M. Murray, ``Consensus and cooperation in
  networked multi-agent systems,'' \emph{Proceedings of the {IEEE}}, vol.~95,
  no.~1, pp. 215--233, 2007.

\bibitem{DeisenrothHH09}
M.~P. Deisenroth, M.~F. Huber, and U.~D. Hanebeck, ``Analytic moment-based
  {Gaussian} process filtering,'' in \emph{Proceedings of the 26th Annual
  International Conference on Machine Learning, {ICML}}, 2009, pp. 225--232.

\bibitem{DeisenrothTHHR12}
M.~P. Deisenroth, R.~D. Turner, M.~F. Huber, U.~D. Hanebeck, and C.~E.
  Rasmussen, ``Robust filtering and smoothing with {Gaussian} processes,''
  \emph{{IEEE} Trans. Automat. Contr.}, vol.~57, no.~7, pp. 1865--1871, 2012.

\bibitem{PanYTB17}
Y.~Pan, X.~Yan, E.~A. Theodorou, and B.~Boots, ``Prediction under uncertainty
  in sparse spectrum {Gaussian} processes with applications to filtering and
  control,'' in \emph{Proceedings of the 34th International Conference on
  Machine Learning, {ICML}}, 2017, pp. 2760--2768.

\bibitem{KoF11}
J.~Ko and D.~Fox, ``Learning {GP}-bayesfilters via {Gaussian} process latent
  variable models,'' \emph{Auton. Robots}, vol.~30, no.~1, pp. 3--23, 2011.

\bibitem{ZhuM10}
M.~Zhu and S.~Mart{\'{\i}}nez, ``Discrete-time dynamic average consensus,''
  \emph{Automatica}, vol.~46, no.~2, pp. 322--329, 2010.

\bibitem{Olfati-Saber07a}
R.~Olfati{-}Saber, ``Distributed {Kalman} filtering for sensor networks,'' in
  \emph{46th {IEEE} Conference on Decision and Control, {CDC}}, 2007, pp.
  5492--5498.

\bibitem{olfati2005distributed}
R.~Olfati-Saber, ``Distributed {Kalman} filter with embedded consensus
  filters,'' in \emph{44th {IEEE} Conference on Decision and Control,
  {CDC}}.\hskip 1em plus 0.5em minus 0.4em\relax IEEE, 2005, pp. 8179--8184.

\bibitem{battistelli2011information}
G.~Battistelli, L.~Chisci, S.~Morrocchi, and F.~Papi, ``An
  information-theoretic approach to distributed state estimation,'' \emph{IFAC
  Proceedings Volumes}, vol.~44, no.~1, pp. 12\,477--12\,482, 2011.

\bibitem{BattistelliCMFG15}
G.~Battistelli, L.~Chisci, G.~Mugnai, A.~Farina, and A.~Graziano,
  ``Consensus-based linear and nonlinear filtering,'' \emph{{IEEE} Trans.
  Automat. Contr.}, vol.~60, no.~5, pp. 1410--1415, 2015.

\bibitem{BattistelliC16}
G.~Battistelli and L.~Chisci, ``Stability of consensus extended {Kalman} filter
  for distributed state estimation,'' \emph{Automatica}, vol.~68, pp. 169--178,
  2016.

\bibitem{PellegriniESG09}
S.~Pellegrini, A.~Ess, K.~Schindler, and L.~J.~V. Gool, ``You'll never walk
  alone: Modeling social behavior for multi-target tracking,'' in \emph{ICCV},
  2009, pp. 261--268.

\bibitem{LuberSTA10}
M.~Luber, J.~A. Stork, G.~D. Tipaldi, and K.~Arras, ``People tracking with
  human motion predictions from social forces,'' in \emph{{IEEE} International
  Conference on Robotics and Automation}, 2010, pp. 464--469.

\bibitem{YamaguchiBOB11}
K.~Yamaguchi, A.~C. Berg, L.~E. Ortiz, and T.~L. Berg, ``Who are you with and
  where are you going?'' in \emph{CVPR}, 2011, pp. 1345--1352.

\bibitem{Helbing95}
D.~Helbing and P.~Moln\'ar, ``Social force model for pedestrian dynamics,''
  \emph{Phys. Rev. E}, vol.~51, pp. 4282--4286, May 1995.

\bibitem{TamuraLHCBYA12}
Y.~Tamura, P.~D. Le, K.~Hitomi, N.~P. Chandrasiri, T.~Bando, A.~Yamashita, and
  H.~Asama, ``Development of pedestrian behavior model taking account of
  intention,'' in \emph{{IEEE/RSJ} International Conference on Intelligent
  Robots and Systems}, 2012, pp. 382--387.

\bibitem{EllisS009}
D.~Ellis, E.~Sommerlade, and I.~D. Reid, ``Modelling pedestrian trajectory
  patterns with {Gaussian} processes,'' in \emph{12th {IEEE} International
  Conference on Computer Vision Workshops, {ICCV} Workshops}, 2009, pp.
  1229--1234.

\bibitem{AlahiGRRLS16}
A.~Alahi, K.~Goel, V.~Ramanathan, A.~Robicquet, F.~Li, and S.~Savarese,
  ``Social {LSTM:} human trajectory prediction in crowded spaces,'' in
  \emph{{IEEE} Conference on Computer Vision and Pattern Recognition, {CVPR}},
  2016, pp. 961--971.

\bibitem{GuptaJFSA18}
A.~Gupta, J.~Johnson, L.~Fei{-}Fei, S.~Savarese, and A.~Alahi, ``Social {GAN:}
  socially acceptable trajectories with generative adversarial networks,'' in
  \emph{{IEEE} Conference on Computer Vision and Pattern Recognition, {CVPR}},
  2018, pp. 2255--2264.

\bibitem{mackay2003information}
D.~J.~C. MacKay, \emph{Information theory, inference and learning
  algorithms}.\hskip 1em plus 0.5em minus 0.4em\relax Cambridge university
  press, 2003.

\bibitem{chen2018priors}
Z.~Chen and B.~Wang, ``How priors of initial hyperparameters affect {Gaussian}
  process regression models,'' \emph{Neurocomputing}, vol. 275, pp. 1702--1710,
  2018.

\bibitem{kheperaIII2013}
F.~Lambercy and J.~Tharin, ``Khepera {III} user manual,'' \emph{K-Team},
  Feburary 2013.

\bibitem{hokuyo2005}
Y.~Mori, ``Scanning laser range finder {URG-04LX} specifications,''
  \emph{Sentek Solution}, October 2005.

\bibitem{EsterKSX96}
M.~Ester, H.~Kriegel, J.~Sander, and X.~Xu, ``A density-based algorithm for
  discovering clusters in large spatial databases with noise,'' in
  \emph{Proceedings of the Second International Conference on Knowledge
  Discovery and Data Mining (KDD)}, 1996, pp. 226--231.

\bibitem{KingmaB14}
D.~P. Kingma and J.~Ba, ``Adam: {A} method for stochastic optimization,'' in
  \emph{International Conference on Learning Representations}, 2015.

\bibitem{5338565}
Y.~{Bar-Shalom}, F.~{Daum}, and J.~{Huang}, ``The probabilistic data
  association filter,'' \emph{IEEE Control Systems Magazine}, vol.~29, no.~6,
  pp. 82--100, 2009.

\bibitem{KamalBFR16}
A.~T. Kamal, J.~H. Bappy, J.~A. Farrell, and A.~K. Roy{-}Chowdhury,
  ``Distributed multi-target tracking and data association in vision
  networks,'' \emph{{IEEE} Trans. Pattern Anal. Mach. Intell.}, vol.~38, no.~7,
  pp. 1397--1410, 2016.

\end{thebibliography}

\end{document}